%% file: shenlong.tex
\documentclass[conference]{IEEEtran}
\IEEEoverridecommandlockouts
\usepackage{cite}
\usepackage{amsmath,amssymb,amsfonts}
\usepackage{algorithmic}
\usepackage{graphicx}
\usepackage{textcomp}
\usepackage{xcolor}
\usepackage{url}

\usepackage[ruled]{algorithm2e} 

\SetAlFnt{\small}
\SetAlCapFnt{\small}
\SetAlCapNameFnt{\small}
\SetAlCapHSkip{0pt}

\usepackage{listings}

\usepackage[most]{tcolorbox}
\usepackage{xcolor}
\usepackage{listings}

\definecolor{background}{HTML}{EEEEEE}
\definecolor{comments}{HTML}{76ABAE}
\definecolor{keywords}{HTML}{31363F}
\definecolor{strings}{HTML}{76ABAE}
\definecolor{highlight}{HTML}{222831}

\lstdefinelanguage{Markdown}{
  morekeywords={__},  
  morecomment=[l]{//},  
  morestring=[b]"",     
}

\lstdefinestyle{mystyle}{
  backgroundcolor=\color{background},
  commentstyle=\color{comments},
  keywordstyle=\color{strings},
  numberstyle=\tiny\color{keywords},
  stringstyle=\color{strings},
  basicstyle=\ttfamily\footnotesize,
  breakatwhitespace=false,
  breaklines=true,
  captionpos=b,
  keepspaces=true,
  numbersep=5pt,
  showspaces=false,
  showstringspaces=false,
  showtabs=false,
  tabsize=2,
  frame=single,
  frameround=ffff,
  morekeywords={MUST, LEFT, RIGHT, BOTTOM, TOP, BACK, FORWARD, APPLY, DO, NOT, USE, ONLY, OUTPUT, FOR, INSTANCE, CONSIDER, JSON, INPUT, MOVE, AND, IMPORTANT, REVISE, DOCUMENT, ADHERE},
  escapeinside={(*@}{@*)},
  moredelim=[l][\bfseries\color{highlight}]{\#\#\#}, 
  moredelim=[s][\color{strings}\bfseries]{**}{**}, 
}

\newtcolorbox{codeblock}[2][]{
  colback=red!5!white,
  colframe=red!75!black,
  fonttitle=\bfseries,
  title=#2,
  arc=5mm, 
  outer arc=5mm, 
  boxsep=5pt, 
  left=5pt, 
  right=5pt, 
  #1
}

\def\BibTeX{{\rm B\kern-.05em{\sc i\kern-.025em b}\kern-.08em
    T\kern-.1667em\lower.7ex\hbox{E}\kern-.125emX}}
\begin{document}

\title{Geometric Algebra Meets Large Language Models: Instruction-Based Transformations of Separate Meshes in 3D, Interactive and Controllable Scenes
}

\author{\IEEEauthorblockN{Prodromos Kolyvakis}
\IEEEauthorblockA{\textit{ORamaVR} \\
Geneva, Switzerland\\
prodromos.kolyvakis@oramavr.com}
\and
\IEEEauthorblockN{Manos Kamarianakis}
\IEEEauthorblockA{\textit{ORamaVR} \\
Heraklion, Greece \\
manos.kamarianakis@oramavr.com}
\and
\IEEEauthorblockN{George Papagiannakis}
\IEEEauthorblockA{\textit{ORamaVR} \\
Geneva, Switzerland \\
george.papagiannakis@oramavr.com}
}

\maketitle

\begin{abstract}
This paper introduces a novel integration of Large Language Models
(LLMs) with Conformal Geometric Algebra (CGA) to revolutionize
controllable 3D scene editing, particularly for object repositioning
tasks, which traditionally requires intricate manual processes and
specialized expertise.
These conventional methods typically suffer from reliance on large
training datasets or lack a formalized language for precise edits.
Utilizing CGA as a robust formal language, our system,
\textit{Shenlong}, precisely models spatial transformations necessary
for accurate object repositioning. Leveraging the zero-shot learning
capabilities of pre-trained LLMs, Shenlong translates natural language
instructions into CGA operations which are then applied to the scene, facilitating exact spatial
transformations within 3D scenes without the need for specialized
pre-training.
Implemented in a realistic simulation environment, Shenlong ensures
compatibility with existing graphics pipelines.
To accurately assess the impact of CGA, we benchmark  against robust Euclidean Space 
baselines, evaluating both latency and accuracy. 
Comparative performance evaluations indicate that Shenlong
significantly reduces LLM response times by 16\% and boosts success
rates by 9.6\% on average compared to the traditional methods.
Notably, Shenlong achieves a 100\% perfect success rate in common practical
queries, a benchmark where other systems fall short.
These advancements underscore Shenlong's potential to democratize 3D
scene editing, enhancing accessibility and fostering innovation across
sectors such as education, digital entertainment, and virtual reality.
\end{abstract}

\begin{IEEEkeywords}
Large Language Models, Conformal Geometric Algebra, 3D Scene Editing, Separate Meshes, Object Repositioning, Controllable Scene Authoring
\end{IEEEkeywords}

\input{body}

\end{document}

%% file: body.tex
\maketitle

\section{Introduction}

\begin{figure*}
\includegraphics[width=0.9\textwidth]{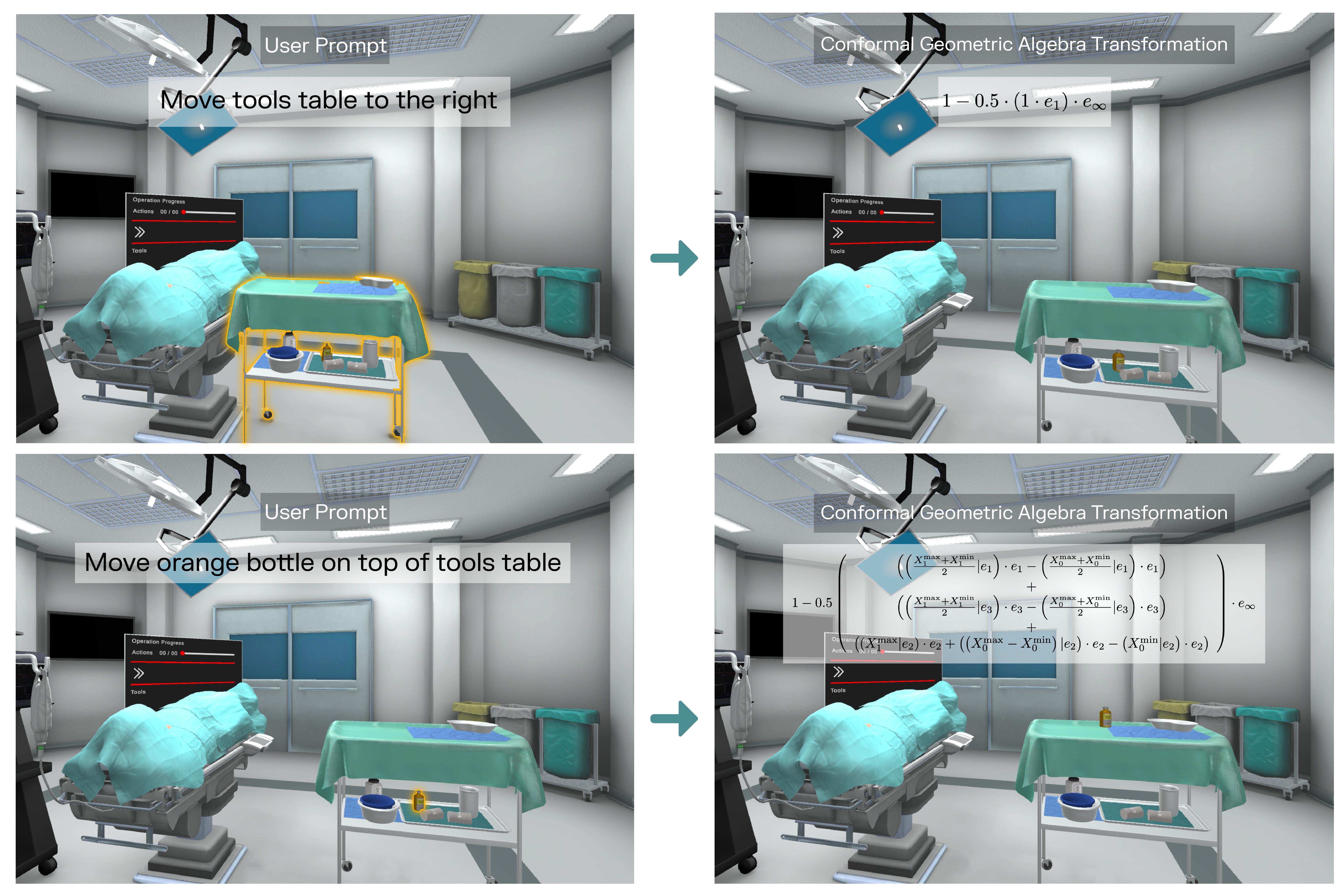}
\centering
\caption{Illustrative demonstration of \emph{Shenlong}, our text-based AI assistant for precise object repositioning in scene edits. Displayed are two edit prompts: on top, \textit{moving the tool table to the right}; and on the bottom, \textit{placing the orange bottle on top of the tool table}. The left side of each panel shows the scene before the edit, and the right side shows the scene after the edit. Shenlong processes user text queries and generates the corresponding transformations in terms of  Conformal Geometric Algebra (CGA), by harnessing a Large Language Model (LLM). The resulting output is then applied to the scene. For each scene, the user's prompt and the LLM's CGA-formatted response are presented. Using CGA instead of baseline representations such as transformation matrices, yields better results as shown in Section~\ref{sub:benchmarking}.}
\end{figure*}

Our primary objective is to enable the instruction-based repositioning of objects within 3D, interactive and controllable scenes.
Such rearrangement tasks are defined through various descriptors such as object poses, visual representations, descriptive language, or immersive interactions by agents within the desired end states.
In this work, we focus primarily on textual instructions.
This activity forms a part of the broader challenge in Embodied AI known as the \textit{rearrangement of environments}, where the goal is to configure spaces into predetermined states \cite{batra2020rearrangement}.
The ability to reposition objects accurately and intuitively is essential, especially given its significant implications across fields such as digital entertainment, virtual reality (VR), simulation training, and architectural visualization.

Traditionally, this task has demanded intensive manual effort and specialized knowledge, restricting the efficiency and accessibility of 3D scene editing.
Recent developments in Foundational Models \cite{Bommasani2021FoundationModels} represent a paradigm shift, suggesting that complex scene editing could become more intuitive and accessible through simple text-based instructions.
However, existing machine learning-based scene representation techniques such as Neural Radiance Fields (NeRFs) \cite{Mildenhall20eccv_nerf} and Gaussian Splatting \cite{kerbl3Dgaussians} present substantial challenges for precise object repositioning due to their holistic nature, which often obscures individual object details and limits model generalization due to the plurality of possible scenes. \cite{Yu20arxiv_pixelNeRF}.
In contrast, using separate mesh scene representations provides a viable solution by granting direct access to individual mesh components, which facilitates intuitive interactions and precise alignments tailored to specific editing requirements \cite{echoscene}.
However, the overarching question remains: \textit{How effectively can machine learning interact with these separate mesh components for editing?}

The advent of Large Language Models (LLMs) \cite{pmlr-v139-ramesh21a,touvron2023llama,jiang2023mistral} offers substantial advancements in human-computer interaction by utilizing linguistic proxies to facilitate instruction-based applications \cite{hong2023metagpt,delatorre2024llmr,gong2024title,durante2024interactive,yang2023holodeck}.
Moreover, LLMs have significantly contributed to the field of Neurosymbolic AI \cite{10148662,garcez_neurosymbolic_2023}, enhancing our capability to transform linguistic instructions into precise symbolic representations that bridge the gap between LLMs' comprehensive understanding abilities and the precision required for complex spatial transformations \cite{NEURIPS2023_deb3c281,jignasu2023foundational,hong2023dllm}.
This LLM - Neurosymbolic AI integration prompts a critical inquiry into the optimal symbolic notation
that encapsulates geometric transformations in a manner that LLMs can
readily interpret. Geometric Algebra (GA), also known as Clifford
Algebra, provides a robust mathematical structure ideal for managing
transformations and interactions of geometric objects, which has been
widely applied in Computer Graphics
\cite{papagiannakis2013geometric,papaefthymiou2016inclusive,siggraphGAcourse2019,siggraphGAcourse2022}. 

In this paper, we use a specific GA model
called \textit{3D Conformal Geometric Algebra} (CGA) to effectively
integrate intuitive linguistic instructions with precise geometric
operations, offering a more accessible method for editing separate
mesh 3D scenes. 
This method utilizes the zero-shot learning capabilities of Large
Language Models (LLMs), which allows for adaptability to new 3D
environments without the requirement for scene-specific training. Our
principal contribution is the development of
Shenlong, a system tailored for separate mesh scene editing,
particularly designed for object repositioning through textual
descriptions. Integrated within the ThreeDWorld (TDW) framework
\cite{gan2021threedworld}, Shenlong supports the Unity3D
Engine and enhances VR-ready 3D scene interaction capabilities.
Through rigorously designed experiments, we show that
Shenlong markedly surpasses existing LLM-based alternatives,
including those used in NVIDIA's Omniverse, in terms of object
repositioning within 3D and interactive scenes. A critical aspect of
our work is the detailed examination of the limitations currently
faced by LLM solutions in object repositioning tasks. Furthermore, we
illustrate that LLMs can proficiently generate and manipulate CGA
operations with minimal input. Ultimately, our research marks a
significant advancement towards democratizing the creation and
manipulation of digital environments by simplifying the complex
technicalities traditionally involved in 3D scene editing.

\section{Related Work}

We now present the background pertinent to our research, outlining the foundational concepts and related advances in the field.

\subsection{Agent-Driven Object Rearrangement}

In the realm of object rearrangement, understanding reusable abstractions through geometric goal specifications ranges from simple coordinate transformations to complex multi-object scenarios \cite{batra2020rearrangement}.
\cite{chang2023hierarchical} approach rearrangement as an offline goal-conditioned reinforcement learning problem where actions rearrange objects from an initial setup in an input image to conform to a goal image's criteria.
\cite{kapelyukh2023scenescore} propose learning a cost function through an energy-based model to favor human-like object arrangements.
\cite{pmlr-v205-simeonov23a} address rearrangement tasks using Neural Descriptor Fields \cite{ndfs}, assigning consistent local coordinate frames to task-relevant object parts, localizing these frames on unseen objects, and aligning them through executed actions.
Furthermore, \cite{echoscene} integrate scene graphs with diffusion processes for editable generative models, yet require extensive training and access to the complete scene graph, highlighting the efficiency of our localized transformation approach. 
Diverging from methods that depend on extensive scene-specific training, our approach leverages the generalization capabilities of LLMs to simplify object repositioning tasks.
In contrast to \cite{kwon2024language}, who limit their LLM applications to Euclidean spaces for predicting end-effector poses, and \cite{cook2ltl}, who necessitate numerous and cumbersome predefined predicates for translating cooking instructions into Linear Temporal Logic, our method expoits CGA and decomposes repositioning tasks into a minimal set of primitive transformations.
Unlike the LLMR approach \cite{delatorre2024llmr}, which uses specialized LLMs to generate C\# Unity code for scene creation and editing, our method avoids direct code generation, thereby ensuring precise scene manipulation without the complexities of writing and debugging code.

\subsection{Machine Learning Applications of Geometric Algebra}
The incorporation of
GA into neural computation was initially introduced in
\cite{pearson1994neural}, with subsequent developments introducing
multivector-valued neurons for radial basis function networks
\cite{corrochano1996selforganizing}, multilayer perceptrons (MLPs)
\cite{buchholz2000quaternionic, buchholz2001clifford}, and various
neural network architectures \cite{buchholz2008polarized,
buchholz2008clifford, buchholz2007optimal, bayro2001geometric}.
Currently, GA-based neural networks have been applied in various
domains, including signal processing \cite{buchholz2008polarized},
robotics \cite{BAYROCORROCHANO201872}, partial differential equation
modeling \cite{brandstetter2022clifford}, fluid dynamics
\cite{ruhe2023geometric}, and particle physics
\cite{ruhe2024clifford}. Furthermore, novel GA-based architectures
such as multivector-valued convolutional neural networks (CNNs)
\cite{li2022ga, wang2021rga}, recurrent neural networks
\cite{kuroe2011models, zhu2016global}, and transformer networks
\cite{liu2022geometric, brehmer2024geometric} have been introduced,
demonstrating the versatility and efficacy of GA in enhancing neural
network capabilities for geometrically oriented tasks.
Unlike these approaches, our work does not incorporate geometric algebraic components directly within deep learning architectures.
Instead, we utilize geometric algebra as a communicative mediator between the LLM and our object rearrangement application.
This method emphasizes the use of geometric algebra primarily as a tool to enhance the interaction and translation of complex geometrical tasks into understandable formats for the LLM, fostering more effective problem solving capabilities in practical applications.
Recently, \cite{llm_ga_teach} fine-tuned ChatGPT with a large curated collection of textual documents on Geometric Algebra, aimed at developing customized learning plans for students in diverse fields.
Our work diverges from the approach of fine-tuning LLMs and instead investigates whether LLMs can effectively utilize geometric algebra for object rearrangement tasks with minimal prompting.
\section{Conformal Geometric Algebra}

In his seminal 1872 Erlangen Programme, Felix Klein introduced the
revolutionary idea that geometry is best explained by algebra,
asserting that algebraic structures govern geometric shapes
\cite{HAWKINS1984442}. Building on this, quaternions have resurged
since the late 20th century due to their compact and efficient
representation of spatial rotations. GA further unifies and extends
algebraic systems, including vector algebra, complex numbers, and
quaternions, through the concept of \textit{multivectors}. Unlike
traditional approaches, which treat scalars, vectors, and
higher-dimensional entities separately, GA provides a cohesive
framework to uniformly represent and manipulate these entities.

At the heart of GA is the concept of the
\textit{geometric product}, which generalizes the dot product and the
cross product in Euclidean spaces. Additionally, as all
products can be defined using only the geometric one, we need only use
the latter one along with addition, scalar multiplication, and
conjugation to perform any multivector manipulation.
In the context of this work we will be employing CGA, a 32-dimensional 
extension of quaternions and dual-quaternions \cite{rooney2007william}, where all entities such
as vertices, spheres, planes as well as rotations, translations
and dilations are uniformly expressed as multivectors \cite{kamarianakis2021all}. 

For example, assuming the standard CGA basic elements $\{e_1, e_2, e_3, e_4, e_5\}$, a basis of CGA would consist 
of all 32 ($2^5$) combinations of up to five of them via geometric product, i.e., $\{1, e_i, e_ie_j, e_ie_je_k, e_ie_je_ke_l, e_1e_2e_3e_4e_5 \text{ for } 1\leq i<j<k<l\leq 5\}$. For convenience, we define the vectors $e_{o}=0.5(e_5-e_4)$ and $e_{\infty}=e_4+e_5$, as well as denote products of basic elements using only subscripts, e.g., $e_{ijk}:=e_ie_je_k$. Using this notation, a sphere $s$ centered at  $x = (x_1,x_2,x_3)$ with radius $r$ amounts to the CGA multivector:
\begin{align}
S = x_1e_1+x_2e_2+x_3e_3 + \frac{1}{2}(x_1^2+x_2^2+x_3^2-r^2)e_{\infty}+e_o.
\end{align} 
Notice that (a) setting $r=0$ 
would yield the respective multivector for the point $x$ and that (b)
given $S$ we can extract both $x$ and $r$.  

In this algebra, a translation by $(t_1,t_2,t_3)$ amounts to¨ 
\begin{equation}
T=1-0.5(t_1e_1+t_2e_2+t_3e_3)e_{\infty},
\end{equation}
thus its inverse would be 
\begin{equation}
T^{-1}=1+0.5(t_1e_1+t_2e_2+t_3e_3)e_{\infty},
\end{equation}
As noted in \cite{kamarianakis2021all},  a rotation that would normally be expressed by the 
unit quaternion:
\begin{equation}\label{eq:quat}
q:=a-d\pmb{i}+c\pmb{j}-b\pmb{k}. 
\end{equation}
can be represented by the respective \textit{rotor}: 
\begin{equation}\label{eq:rotor}
R = a+be_{12}+ce_{13}+de_{23}.
\end{equation}
It then becomes apparent that the inverse of $R$ is 
\begin{equation}\label{eq:R_inverse}
R^{-1} = a-be_{12}-ce_{13}-de_{23}.
\end{equation}
Finally, the multivector:
\begin{equation}\label{eq:dilation}
D = 1 + \frac{1-d}{1+d}e_{45}
\end{equation}
corresponds to a dilation of scale factor $d>0$ with 
respect to the origin, whereas it holds that:
\begin{equation}\label{eq:dilation_inverse}
D^{-1} = \frac{(1+d)^2}{4d} + \frac{d^2-1}{4d}e_{45}.
\end{equation}
Equations \eqref{eq:dilation} and \eqref{eq:dilation_inverse} appear in bibliography with $e_{45}$ replaced by the 
equivalent quantity $e_\infty\wedge e_o$, 
where $\wedge$ denotes the \textit{wedge} (or outer) product.

\begin{figure*}[t]
    \includegraphics[width=\linewidth]{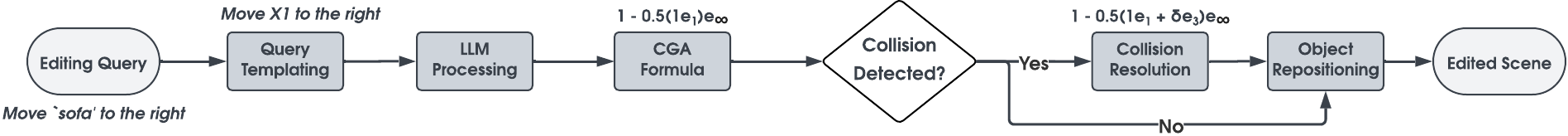}
    \caption{Shenlong's overview: It generates a CGA transformation after \emph{templating} the query and adds a \(\delta\) upward translation for a hypothetical Z-axis collision.}
    \label{fig:overview}
\end{figure*}

To apply transformations $M_1, M_2, \ldots, M_n$ (in this order) 
to an object, we define the multivector $M := M_n M_{n-1} \cdots M_1$, 
where all intermediate products are geometric. Shenlong is capable 
of identifying and generating the intermediate transformations $M_i$ that 
should be applied to an object and properly returns $M$. As a composition of 
rigid body objects and dilations, the resulting 
multivector is known to be equivalent to a simpler product $M=TRD$
of translation $T$, a rotation $R$ and a dilation $D$. By extracting 
$T,R$ and $D$, we can express the encapsulated information of $M$ 
as a translation vector, a unit quaternion and a scale factor which 
can be ingested by TDW and Unity to visualize the edited object.

To extract $T,R$ and $D$ from a generated CGA product $M=TRD$, we follow the 
methodology presented in \cite{GAUnity}. Specifically,
we apply $M$ to a sphere $C$, centered at 
origin, with radius equal to $1$. The multivector $C'$ corresponding to the transformed sphere 
can be evaluated via  the \textit{sandwich product} $C' = MCM^{-1}$, 
and we can therefore extract its center $x$ and its radius $r$.
However since $C'$ is the image of the unit, origin-centered, sphere $C$ after applying $M=TRD$, 
i.e., first a dilation by $D$, then a rotation by $R$ and finally a translation by $T$, its 
radius should be equal to the scaling factor of $D$ and its center should be the translation 
vector corresponding to $T$. Since $T$ and $D$ are therefore known, along with 
$M=TRD$, we can evaluate $R := T^{-1}MD^{-1}$ and therefore the corresponding unit 
quaternion using \eqref{eq:quat} and \eqref{eq:rotor}.

\section{Methodology}

Below we provide the high-level overview of our system, illustrated in Figure~\ref{fig:overview}.
When Shenlong accepts a query, e.g., \emph{move the `sofa' to the right}, it replaces each object 
name with a corresponding variable, such as \emph{move X1 to the right}.
Shenlong accepts object names enclosed in quotes and sequentially assigns new variables 
for each unique match found.
The templated queries generated allow the LLM to perform symbolic reasoning.
Following the substitution of object names with variables, Shenlong prompts an LLM to generate the 
corresponding CGA transformations for the user's query, 

It should be noted that Shenlong operates at a local level, lacking a comprehensive understanding of other objects in the scene.
While this approach enhances processing efficiency, it may result in collisions with other objects.
To address this, we observe that the fuzziness of queries, such as \emph{move the `sofa' to the right}, allows for multiple possible solutions.
Upon detecting a collision, Shenlong explores perturbed  transformations to resolve the issue, e.g., a \(\delta\) upward translation on Z-axis, ensuring minimal deviation from the intended initial transformation.
As a final remark, the proposed system is integrated within the ThreeDWorld (TDW) framework \cite{gan2021threedworld},  supports Unity3D Engine and fosters VR-ready 3D scenes, augmenting human VR interaction.

\subsection{LLM Processing \& CGA Formulation}

Shenlong harnesses CGA to express spatial transformations in a 3D environment, leveraging the precision of this rigorous mathematical framework.
Shenlong replaces object names with variables; a vital step for abstracting the user's intent and preparing the data for algebraic processing.
Each object $X_i$ in the scene is represented by an axis-aligned bounding box defined by two points, $X_i^{\text{max}}$ and $X_i^{\text{min}}$, which encapsulate the object's spatial extent.
This bounding box model simplifies the calculation of movements and rotations by providing clear, definable limits to each object's position.

Through careful prompting, we instruct the LLM on the key facts and operations of CGA.
Below are some key remarks.
We explained how coordinate extraction can utilize the inner product $|$ operation to precisely isolate spatial coordinates.
For example, \((X_1^{\text{max}} | e_2) \) corresponds to the $y$ coordinate
of $X_i^{\text{max}}$.
Additionally, we defined the outer (or wedge) product, which establishes planes for rotational operations, and details the mechanisms for both translation and rotation rotors.
These rotors are not only defined individually but are also combined through rotor composition, enabling complex sequential transformations essential for accurate scene manipulation.
To effectively guide the LLM in utilizing the rotation and translation rotors, we provided a total of five illustrative examples.
These include a single example each of rotation and translation to demonstrate basic movements, one compositional example that integrates both types of transformations, and two additional examples addressing more complex, fuzzy queries, such as \emph{move on top of} and \emph{move next to} another.
This small diverse set of examples ensures the LLM effectively understands and implements the necessary algebraic operations for object repositioning.
Despite the limited number of examples, i.e., only five, we demonstrate in Section \ref{sec:results_discussion} that the LLM can still generalize to complex queries.
Finally, the LLM responds by generating a JSON output that includes the specified rotors for composition to be applied to each object.

\subsection{Collision Detection Module}

Similar to \cite{yang2023holodeck}, our system conducts a Depth-First Search (DFS) 
on a constructed 3D grid surrounding the colliding objects, identifying the first 
valid solution within the considered search space.
We only consider collisions that can be determined using the bounding box information 
for each object. To prevent placing objects directly on the corners of the 3D grid, 
we incorporate a fixed buffer zone to ensure adequate spacing during object placement.
Each object is described by 6 parameters, corresponding to 
its axis-aligned bounding box : \( (x, y, z) \) for the center coordinates, \( w, d \) 
and \( h \) for the width, depth and height.
The process starts with initial placements of the target objects and concludes at 
the first occurrence of a valid placement.
We prioritize solutions that are close to the original configurations, exploring all 
possibilities and gradually allowing for more distant solutions in a linear progression.
Our goal is to minimize disruptions from the initial configuration while preserving the 
spatial integrity of the scene.
This collision resolution operates within a set timeframe (e.g., 0.5 or 1 second).
Finally, all transformations are applied. 

\section{Evaluation}

We conduct comprehensive human evaluations to assess Shenlong in object repositioning, engaging 20 annotators: 5 3D designers, 5 game programmers, and 10 individuals outside the gaming and 3D design discipline areas.
This study examines Shenlong's proficiency in editing diverse scenes.
Through these user studies, we demonstrate that, by harnessing CGA in our prompting strategies, we significantly outperform baseline alternatives, including those made with NVIDIA Omniverse (see 
Section~\ref{sub:baselineSystems}).

\subsection{Experimental Setup}

We generated ten diverse scenes using the HOLODECK framework \cite{yang2023holodeck} for human evaluation, encompassing various settings including living rooms, wine cellars, kitchens, and medical operating rooms.
To facilitate the integration of diverse scenes into the TDW framework, we developed a custom importer.
For each scene, we created five variations of templated queries, populating them with objects randomly selected from within the scene.
We evaluated 50 distinct queries for each scene, resulting in a total of 2,500 scenes (coupled with prompts) for human assessment.
For each query, we compared the initial scene with the resulting scene after processing by the two baseline system and Shenlong.
Examples of the human evaluation are presented in the \emph{Supplementary Material}.
For each edited scene, we displayed top-down view images and a 360-degree video view, asking annotators to assess the accuracy of the editing performed.
Each object repositioning query was evaluated by five annotators, and a result was considered valid only if all annotators unanimously agreed on the assessment.
It should be noted that all preliminary experiments on dilation-related queries with Shenlong and other baselines achieved a perfect success rate, leading us to omit these queries.
The queries are categorized into five groups, each containing ten queries that evaluate the system's performance across a progressively increasing difficulty gradient.
Below, we provide further details on the group categories used in our analysis:
\begin{itemize}
    \item \textbf{Simple Queries}: Cover basic actions such as rotations and movements along specified axes or planes of a single object.
    \item \textbf{Compositional Queries}: Involve combinations of relative movements and rotations among two or more objects.
    \item \textbf{Fuzzy Queries}: Task systems with interpreting and executing spatial and orientation-specific actions, such as proximity adjustments and directional alignments.
    \item \textbf{Compositional Fuzzy Queries}: Combine multiple elements from compositional and fuzzy queries.
    \item \textbf{Hard Queries}: Represent the most challenging scenarios that test the limits of each system's processing capabilities.
\end{itemize}
The classification of ``simple'' and ``composite'' queries is based on the study of \cite{Manesh2024HowPP} on natural language in virtual environment creation.
Building on this, our work further explores fuzzy and more complex scenarios involving the repositioning of multiple objects.

\subsubsection{System Configuration}

We employed OpenAI's GPT-4 model, specifically the \textit{gpt-4-1106-preview} variant, to process scene editing queries.
Shenlong executed a distinct API call for each query.
The evaluation was done on a laptop with Ubuntu 22.04.4 LTS and AMD Ryzen™ 9 4900HS Mobile Processor (8-core/16-thread, 12MB Cache, 4.3 GHz max boost), NVIDIA® GeForce RTX™ 2060 with Max-Q Design 6GB GDDR6 and 16GB DDR4 RAM at 3200MHz. Rendering was performed using Unity™ 2022.3.9f1.

\subsection{Benchmarking}
\label{sub:benchmarking}

In this section, we assess the performance of our system compared to established baselines.
In the following, we provide an overview of the baseline systems against which our solution is measured, the performance metrics that form the criteria for comparison, and a detailed discussion of the results.

\subsubsection{Baseline Systems}
\label{sub:baselineSystems}

One of our baseline systems, for comparison is one of the publicly available prompts from NVIDIA's Omniverse\footnote{\url{https://github.com/NVIDIA-Omniverse/kit-extension-sample-airoomgenerator/blob/main/exts/omni.example.airoomgenerator/omni/example/airoomgenerator/prompts.py}}.
This prompt operates by generating a JSON output that details object placements within a 3D space, similar to our system.
Contrary to Shenlong, the Omniverse prompt imposes no constraints on the reasoning scheme that the LLM should follow, requesting only the final positions of the objects.
The Omniverse prompt accepts input specifications for each object, including its name, dimensions along the $X$, $Y$ and $Z$ axes, and a centrally located origin point.
To ensure this baseline prompt is comparable to our system, we include only points of interest relevant to the scene editing query, rather than the entire scene description as initially done.
This refined input approach significantly impacts the LLM's response time.
Our experiments demonstrate that providing only the relevant points of interest leads to a substantial decrease in response time—specifically, an average of \(3.3 \pm 0.1\) times faster—without compromising overall accuracy.
Due to these findings, we report results exclusively from this optimized methodology.
Finally, we have extended the prompt by incorporating Euler angles for object orientation.

\begin{figure*}[htbp]
  \centering
  \begin{minipage}[b]{0.48\linewidth}
    \includegraphics[width=\linewidth, height=140pt]{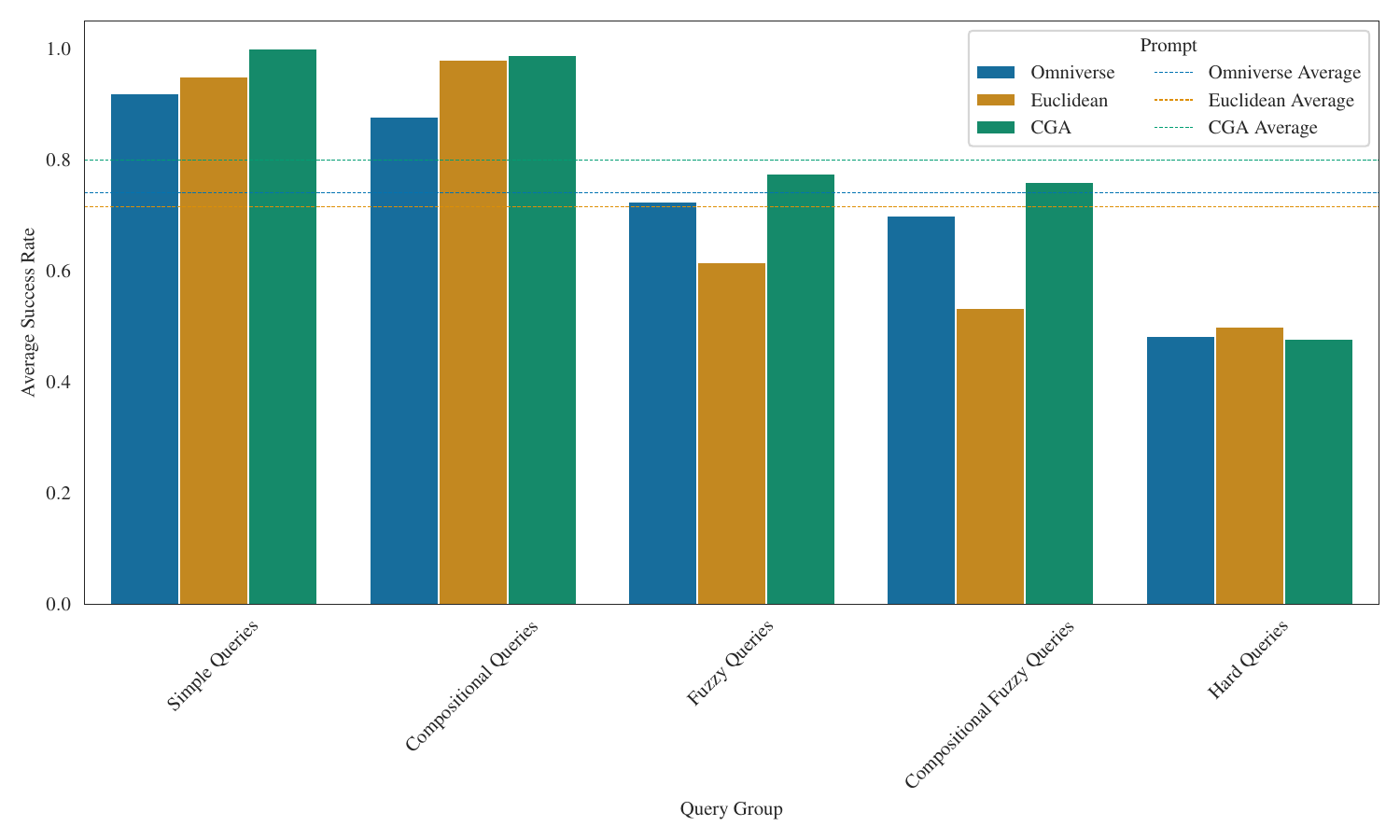}
  \end{minipage}
  \hfill
  \begin{minipage}[b]{0.48\linewidth}
    \includegraphics[width=\linewidth, height=140pt]{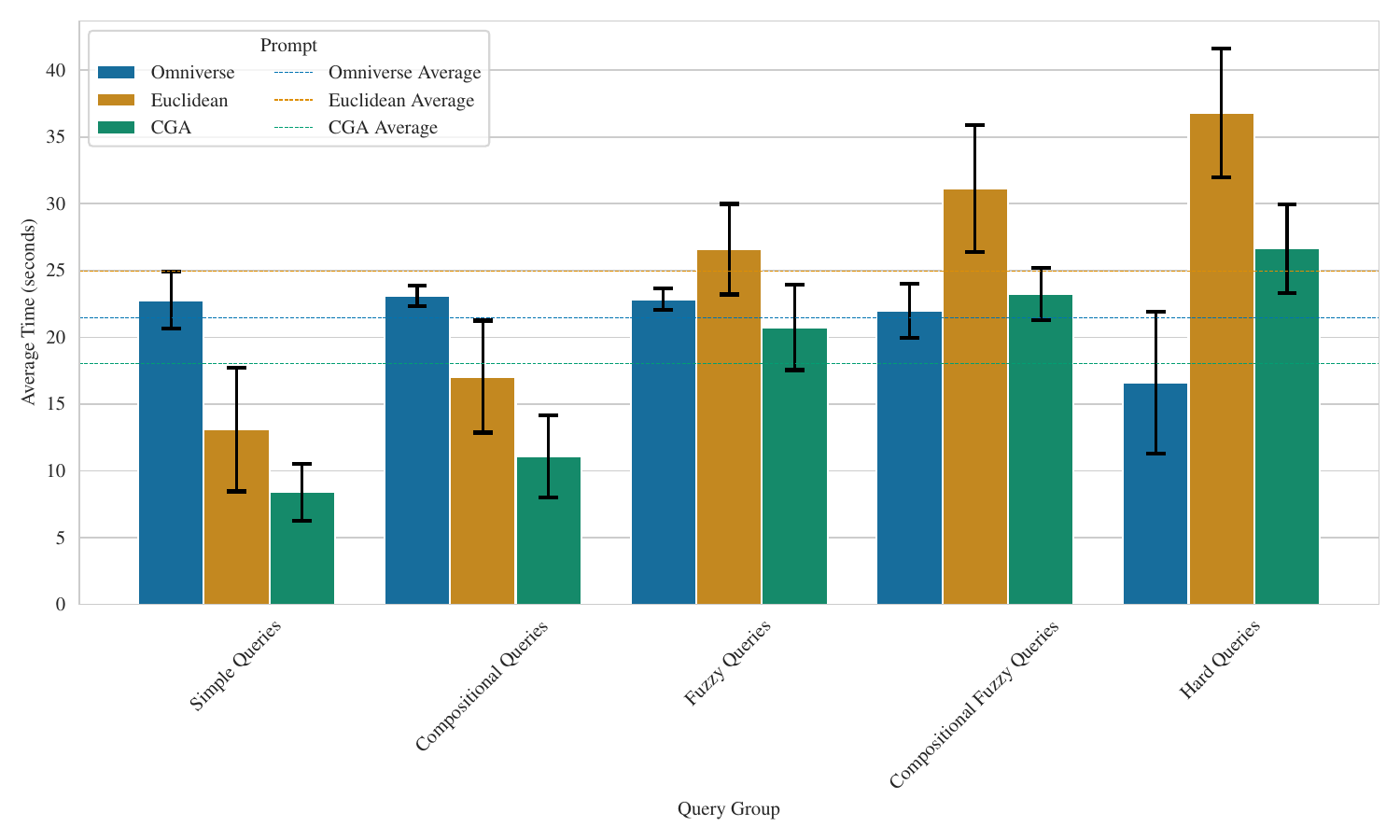}
  \end{minipage}
  \caption{The left chart compares the average success rates, and the right chart compares the average LLM response times for three different systems—Omniverse, Euclidean, and CGA—across five categories of queries: Simple Queries, Compositional Queries, Fuzzy Queries, Compositional Fuzzy Queries, and Hard Queries. Success rates are measured on a scale from 0.0 to 1.0. Horizontal dashed lines indicate the overall mean success rate and response times for each system.}
  \label{fig:grouped_performance}
\end{figure*}

Our second baseline model is derived from recent work showing that LLMs can predict a dense sequence of end-effector poses for manipulation tasks \cite{kwon2024language}.
We extend this concept to object repositioning on 3D scenes, treating it as a form of zero-shot trajectory generation.
In this context, we task the LLM with constructing the necessary translation and rotation matrices to define the final trajectory, adapting the underlying techniques to suit scene manipulation objectives.
We consider this baseline model to be the closest to our approach, as it guides the LLM to operate using templated functions that correspond to translation and rotation functions within the Euclidean space.
This method aligns with our use of structured prompts that direct the LLM to generate specific geometric transformations necessary for scene editing tasks.

To ensure a fair comparison, we append five specific examples to both the baseline prompts and the final guidelines (refer to Section \textit{Additional Guidance for LLM} in Figure~\ref{fig:cga_prompt} ).
This inclusion is based on findings that demonstrated significant benefits in two distinct cases, enhancing the effectiveness of the LLM in performing scene editing tasks.
Due to space limitations, we have included the exact baseline prompts in the \emph{Supplementary Material}.

\subsubsection{Performance Metrics}

To evaluate the system's performance, we calculated the success rate per query across the ten different scenes, each with five templated variations. 
The success rate \( S \) for each scene editing query is calculated as follows:
\begin{equation}
S = \frac{1}{M} \sum_{i=1}^{M} \chi_{ \text{correct}}(i)
\end{equation}
where \( \chi_{ \text{correct}}(i) \) is the characteristic function that equals 1 if the editing query was performed correctly for the \( i \)-th query, and 0 otherwise.
Here, \( M \) represents the total number of queries, calculated as \( M = N \times k \), where \( N \) is the number of different scenes considered (10 in our case), and \( k \) is the number of variations per scene (5 in our case).
Similarly, we compare the average response time across the ten different scenes and their respective variations. The average response time \( T \) is calculated by:
\(
T = \frac{1}{M} \sum_{i=1}^{M} t_i
\), where \( t_i \) represents the response time for the \(i\)-th query.
It is commonly noted that LLMs may not always produce valid outputs.
We adhere to the standard practice of allowing up to \(n\) retries.
The reported timings include these retry durations. Notably, in all our experiments, LLMs generated valid prompts within \(n = 5\) retries.

\subsubsection{Results \& Discussion}
\label{sec:results_discussion}

In this section we present a comprehensive analysis that spans overall performance metrics, detailed evaluations by query group, and granular analyses of individual query performances.
These discussions aim to highlight significant findings, interpret the implications of the results, and explore potential areas for further improvement.
In all reported figures, we refer to the modified baseline prompts for scene editing as \emph{Omniverse} and \emph{Euclidean}, respectively derived from NVIDIA's Omniverse usage examples and methods akin to zero-shot trajectory generation. We name the latter \emph{Euclidean} because it closely aligns with our method's approach in guiding the LLM to generate transformations such as rotations and translations within Euclidean space, in contrast to our use of CGA.
Finally, we refer to Shenlong's prompt as \emph{CGA}.

\begin{figure*}
\begin{lstlisting}
### Objective:
You MUST calculate the appropriate transformation for objects in a 3D space using templated variables (`X1`, `X2`, etc.). 
You MUST only use a valid combination of the defined algebraic operations.
### Coordinate System:
- **X-axis (`e1`)**: Translates the object LEFT (-) or RIGHT (+).
- **Y-axis (`e2`)**: Translates the object BOTTOM (-) or TOP (+) height wise.
- **Z-axis (`e3`)**: Translates the object BACK (-) or FORWARD (+).
### Defined Algebraic Operations:
1. **Rotation** The rotation rotor R(theta, u, v) rotates a vector by an angle theta in the plane defined by u ^ v.
2. **Translation** The translation rotor T(x * e1 + y * e2 + z * e3) translates an object along the vector x * e1 + y 
   * e2 + z * e3.
3. **Coordinate Extraction** Use the inner product (x1 | e3) * e3 to extract the coordinate along the Y-axis.
3. **Outer Product** For independent vectors, u ^ v is non-zero and represents an oriented plane.
4. **Commutation:** Translation operators commute, `T(v) * T(w) = T(w) * T(v)`.
5. **Rotor Composition:** Rotors can be combined by multiplication, representing sequential rotations.
### Task Instructions:
1. APPLY the specified T for translation and R for rotation operations, adhering to the defined algebraic operations.
2. DO NOT introduce any variables or constants not explicitly mentioned in the input.
3. USE ONLY the variables names and e1, e2, e3 on the operations.
4. OUTPUT the transformations in JSON format, using key-value pairs for each object and its transformation.
5. USE ONLY the max and min points of an object's bounding box (e.g., X1.max and X1.min) for all calculations.
6. For coordinate extraction, employ the inner product operator |. For instance, use (X1.max | e2) * e2 to extract the Y-axis coordinate.
7. CONSIDER the current position of objects when moving one in relation to another. Calculate transformations using the initial locations derived from their bounding box points.
### Example Inputs, Results in JSON format
1. **Input:** 'Translate X1 by 1 unit to the left.'
   **Output:** {'X1': 'T(-1 * e1)'}
   **Explanation:** The bounding box of X1 is defined by the points X1.max and X1.min. To translate X1 by 1 unit to the left, we must subtract 1 from the X-axis coordinate of both points. Therefore, the transformation is T(-1 * e1).  
2. **Input:** 'Rotate X1 by 90 degrees around the z-axis.'
   **Output:** {'X1': 'R(np.pi/2, e1, e2)'}
   **Explanation:** The rotation rotor R(theta, u, v) rotates a vector by an angle theta in the plane defined by u ^ v. In this case, the rotation is 90 degrees in the z-axis is defined as the rotation around e1, e2 plane. Therefore, the transformation is R(np.pi/2, e1, e2).
### Output Format:
The output MUST be formatted as a SINGLE JSON object containing all transformations for each object.
### Additional Guidance for LLM:
- ADHERE strictly to the defined algebraic operations.
- DOCUMENT Each Step: For every action you perform, you MUST provide a clear and   numbered list of operations used during your reasoning.
- REVISE any transformation that does not comply with these principles.
- It's **IMPORTANT** to take into consideration all axis when performing relative transformations. Focus on the X **AND** Z axis when moving objects around the room. Make changes in the Y axis when a vertical movement is required. So, for example, when moving an object next to another, you should move it in both the X and Z axis, and when moving an object on top of another, you should move it in the X and Z axis, and also in the Y axis.
- When no axis is defined for rotations, the default axis is the Y-axis which means a rotation in the e1, e3 planes.
- When you want to move or rotate an object along an axis you must use the inner product operator |. For example, to move an object along the Y-axis you must use (X1.max | e2) * e2.
- When no unit is specified for the translation, the default unit is 1 and for the rotation, the default unit is 90 degrees.
\end{lstlisting}
    \caption{Template for Conformal Geometric Algebra Prompts used by Shenlong.}
    \label{fig:cga_prompt}
\end{figure*}

\textbf{Performance Analysis by Query Group}
Figure~\ref{fig:grouped_performance} presents a detailed evaluation of the prompts' performance across different groups of queries.
Each group's results are analyzed to highlight specific strengths and weaknesses of the system in handling varying complexities.
Specifically, the effectiveness and efficiency of the CGA, Omniverse, and Euclidean prompts were evaluated across the five categories of queries: Simple, Compositional, Fuzzy, Compositional Fuzzy, and Hard.
We focused on two key performance metrics: average success rate and average response time, providing insights into each system's scene editing capabilities.

Overall, CGA achieves the highest average success rate of 0.80 and the lowest average response time of 18 seconds (p $<$ 0.05), with the results being statistically significant.
Interestingly, all prompts exhibited the same overall standard deviation, both in terms of success rate and response time, indicating consistent variability across the different editors.
To assess the differences between the systems, we performed Student's t-tests.
In comparison, the Omniverse and Euclidean prompts achieve success rates of 0.74 and 0.72, and average response times of 21.5 seconds and 24.5 seconds, respectively.
While the differences in success rates among Omniverse and Euclidean prompts are not statistically significant, the variations in average response times are.
We conjecture that the lack of statistical significance in terms of success rates among the systems can be attributed to the nature of the reasoning methods employed.
Both the Omniverse and Euclidean prompts rely on classical geometrical reasoning, which may lead to similar levels of performance.
In contrast, Shenlong utilizes a different approach to reasoning, which not only reflects in its superior performance but also in its statistical significance.
This suggests that the distinct reasoning method employed by our system may contribute to its enhanced effectiveness.

Focusing on the success rate metric, CGA consistently exhibits high success rates across all query types, particularly excelling in complex scenarios such as Compositional Fuzzy queries, indicating robust scene understanding.
Given that the success rates of the Euclidean and Omniverse systems are not statistically significant, it is evident that CGA provides a relative boost of 9.6\% over their average performance.
In contrast, Omniverse demonstrates variable performance, experiencing a drop in simpler queries but excelling in fuzzy queries.
Conversely, the Euclidean system performs well in simpler queries but shows significant weaknesses in both fuzzy and composite fuzzy queries.
CGA appears to integrate the advantages of both systems, performing exceptionally well across simple, fuzzy, and compositional queries.
However, it is important to note that with all prompts, ChatGPT-4 shows average performance on hard queries, suggesting potential deficiencies in spatial reasoning.

Regarding response time, Omniverse exhibits consistent average response times across various query groups, regardless of query difficulty, with the notable exception of hard queries, which show the slowest response times.
Importantly, CGA demonstrates a 16\% relative decrease in response time compared to Omniverse, which is the fastest among the baseline systems.
In contrast, the Euclidean prompt demonstrates an increasing trend in response times correlating with query difficulty, peaking with fuzzy, compositional fuzzy, and hard queries.
This trend is noteworthy as it suggests that response times increase as query complexity rises.
The Euclidean prompt consistently records the longest response times across the complex queries, posing challenges in time-sensitive applications.
Meanwhile, the CGA prompt also displays an increasing trend in response times with escalating query difficulty.
Notably, it presents slower response times for simple and simple compositional queries despite achieving a 100\% success rate.
Overall, CGA maintains competitive response times suitable for practical applications, with its response times scaling appropriately with the complexity of queries and achieving the best overall response performance.

\section{Limitations \& Future Work}

\textbf{Object Representations}
Currently, our system utilizes multiple intermediate representations of objects, including textual, templated, and bounding box references, requiring exact object names.
To enhance this, we plan to implement semantic similarity measures at the token level or more advanced similarity searches using distributed representations.
Additionally, we aim to enrich the system with further object information, such as orientation information, to better handle sophisticated queries involving relative rotations.
By improving the system’s grasp of spatial relationships, improved performance in complex scenarios is expected.

\textbf{System Optimisations}
Since we preprocess and create templated queries, it is straightforward to perform query caching optimizations \cite{cook2ltl}.
We aim to implement query caching optimisations to enhance both the accuracy and responsiveness of our system. 
We also envision transforming the collision module into a multimodal agent \cite{yang2023holodeck}, enabling nuanced handling of complex scenarios.
The agent could be provided with a top-down view of the scene to facilitate finding better resolutions.
Lastly, we plan to integrate speech interaction capabilities into our system.

\textbf{Spatial Reasoning}
Our work sheds more light on the spatial reasoning capabilities and limitations of LLMs. Although better and curated prompting could improve performance on more challenging queries, we conjecture that more specialized agents are needed.
Future work will explore multimodal alternatives that incorporate top-down views of the scene along with linguistic queries or LLMs fine-tuned on GA \cite{llm_ga_teach}.

\section*{Acknowledgments}
This work was partially funded by the Innosuisse Swiss Accelerator (2155012933-OMEN-E). We would like to thank Dimitris Angelis for his valuable contribution in this work. 

\bibliographystyle{plain}
\bibliography{references}